# Recurrent neural network based decision support system

Abiodun Ayodeji[*]  Yong-kuo Liu

Fundamental Science on Nuclear Safety and Simulation Technology, Harbin Engineering University, Harbin, Heilongjiang 150001, China.

**Abstract**
Decision Support Systems (DSS) in complex installations play a crucial role in assisting operators in decision making during abnormal transients and process disturbances, by actively displaying the status of the system and recording events, time of occurrence and suggesting relevant actions. The complexity and dynamics of complex systems require a careful selection of suitable neural network architecture, so as to improve diagnostic accuracy. In this work, we present a technique to develop a fault diagnostic decision support using recurrent neural network and Principal Component Analysis (PCA). We utilized the PCA method for noise filtering in the pre-diagnostic stage, and evaluate the predictive capability of radial basis recurrent network on a representative data derived from the simulation of a pressurized nuclear reactor. The process was validated using data from different fault scenarios, and the fault signatures were used as the input. The predictive outputs required are the location and sizes of the faults. The result shows that the radial basis network gives accurate predictions. Selected hyperparameters and diagnostic results are also presented in this paper.

Keywords: Fault diagnosis, recurrent neural network, operator support system, neural network.

(*E-mail: abiodun.ayodeji@hrbeu.edu.cn)

## 1. Introduction

Decision support for system fault diagnostic and isolation (FDI) as applied in complex industrial systems such as nuclear power plants (NPPs) has many facets. One of the most important facets is the early detection of small, incipient faults in plant's components and during transients. The system presents necessary and sufficient information on time to the plant operator to take appropriate actions to mitigate abnormal situation. The classical first principle fault diagnosis approaches for nonlinear complex dynamic system like NPP has a number of limitations. One, its applications have been limited to highly abstracted demonstrations. Another is the presence of large mismatch and uncertainties in the linearization of the non-linear model that does not satisfactorily represent the physical plant. Moreover, on the account of the nuclear plants' inherent complexity, it is difficult to establish a quantitative physical model of nuclear power plant. Besides, it is known that NPP possesses many transient states, as a result of its complexity. Modeling these states using state space representation is quite difficult, and the data-driven methodology provides speedy means of fault detection.

Increasingly, application of neural networks and its variants for FDI has moved from toy examples to real-world systems. Recently, the diagnostic tasks in decision support systems (DSS) are being performed with the help of Artificial Neural Networks (ANNs), by operating on a large knowledge base, which is developed by collecting the plant's time-dependent transient data from the system. Improvements in the application of ANN for identification of dynamic events is as a result of a number of reasons: the ANN's characteristics of adaptive learning, generalization ability, fault tolerance, robustness to noisy data, and parallel processing ability (Roverso, 2000).

In analyzing a complex system using ANN, the main concern for diagnostic engineer is how to choose appropriate network architecture and (non-linear) function to represent the overall output from the neurons.  Besides, for every learning model there is a data distribution on which it will fare poorly (Ayodeji et al. 2018). Many functions and methods for selecting suitable network architecture have been researched and well documented in the literature. However, there is no consensus on any single approach, and it is still widely believed that network selection result from experimental trial and error. Meanwhile, some of the widely use data-driven approaches are the Gaussian radial basis neural network, applied to the problem of identifying nuclear accidents in a Pressurized Water Reactor (PWR) nuclear power plant (Venkat et al. 2003). A distributed fault diagnostic strategy with BP neural network has been used for local diagnosis for subsystems (Gomes et al., 2015). Also, Gottlieb et al., (2006) presented a study on transient identification using support vector machines in nuclear power plants. Principal Component Analysis (PCA) methodology has also been used for pattern recognition using a case study of Bruce B zone-control level oscillations in CANDU reactors by Nasimi, et al., (2014). Jun et al., (2014) developed a fault diagnosis expert system for marine nuclear power plants using multi-layer flow model incorporated with an expert system. LIND et al., (2014) studied the application of multi-layer flow model in a PWR primary system and a primary heat transfer system of the Fast Breeder Reactor (FBR).

Following the success of the 300MW Qinshan I NPP full scope simulator, this research serves as a necessary foundation for building a comprehensive knowledge base for the advanced Qinshan II NPP operator support system. This work, based our previous publication (Ayodeji et al., 2018), presents the developmental process for decision support system. Here, we determined the optimum recurrent neural network architecture necessary to build a comprehensive knowledge base for the decision support system applicable in complex industrial installations. This work compares the predictive result of a number of RBFN neural network architectures for predicting the severity and location of incipient faults in different parts of the plant. As a pilot study, a spectrum of break sizes and incipient fault in different locations are analyzed to assess the performance of the neural network architectures.

This work is arranged as follows: the first section provides the theoretical background for this research. The second section explains the methodology while the third section presents our experimental results. In the fourth section, we discussed the result and conclude the research work.

## 2. Methodology

### 2.1 Theory of the decision support system

The proposed method uses the hybrid of Principal Component Analysis (PCA) and Artificial Neural Networks (ANN) to support fault diagnostic decisions. In the application of ANN for diagnostic tasks, the choice of the input sequence is of paramount importance. Besides, the specific characteristics of the data from NPP - such as high measurement noise resulting from sensor drift and radiation effects on measuring equipment - could affect the modeling accuracy and generalization ability of ANN. To estimate the predictive ability of RBFN to a specific representative data from NPP, two different fault scenarios were considered in this work, and the model performance was evaluated, based on the data obtained from the simulation of Qinshan II NPP. In this research, PCA was applied to the input data, for feature extraction, dimensionality reduction, noise filtering and data cleaning. The purpose is to select the most informative predictive variable sequences (principal components). Then we experiment with Radial Basis Function Network (RBFN), to determine the generalization capability and the network architecture suitable for the task of fault identification on a representative data from simulated Qinshan II NPP.

### 2.2. Feature selection with PCA

NPP operations are characterize by disturbances, noise in measurement and state variations, and a number of false alarm is generated as a result of these transient change in states. Modeling uncertainties such as model abstractions, disturbances and high background noise as found in NPPs may not be critical to the process behavior but can obscure fault detection by raising a false alarm (Patton et al., 2000). To compensate for the model uncertainties as a result of fluctuations and instability in the operation of NPP, to improve regression accuracy of ANN and for optimization purposes, PCA is first applied to the input vectors.

Selecting the principal components involves the elimination of components that gives the least information about the representative data. In this work, PCA is used to reduce the dimension of the plant variables into a few relevant variables that represents the parameters with significant changes or principal components, which serves to eliminate redundant components. Mathematically, orthogonalizing the components of network inputs can be expressed as follows: A data matrix X, with $m$ number of samples and $n$ number of variables can be decomposed into its principal components as:

$$X = t_1 P_1^T + t_2 P_2^T + .... + t_l P_l^T + t_m P_m^T \tag{1}$$

Where $t$ is defined as the score vectors, and $Pi's$ are the transposed load vectors. Hence, when the $t$'s are derived and arranged in descending order, eliminating the lower value $t$'s means eliminating the components with low variance. Alternatively, the decomposition may be of the form:

$$X = \hat{X} + E \tag{2}$$

Where $\hat{X}$ is the required value of $X$, and $E$ is the uncertainty in the model, both of which are defined as follows:

$$\hat{X} = \hat{T}\hat{P}^T = \sum_{i=1}^{l} t_i p_i^T \tag{3}$$

$$E = \tilde{T}\tilde{P}^T = \sum_{i=l+1}^{m} t_i p_i^T \tag{4}$$

We obtained the required zero-mean $\hat{X}$ by normalizing the original input vector X, and we implemented the MATLAB 'processpca' on the input data to eliminate those components that contribute less than 2% to the total variation in the data set.

### 2.3. Fault isolation and estimation

The radial basis function network is an implementation of neural network that utilizes a single hidden layer to map a $D$-dimensional non-linear input to a $J$-dimensional space of functions in the hidden layer. The general structure of RBFN consists of two layers: a hidden layer which performs a non-linear transformation from input space to hidden space, and the output layer that supplies the response. The source node starts from the input and propagates to the hidden layer. The hidden layer has a Gaussian (radial) transfer function and the output layer has linear activation function. The weights and biases of each neuron in the hidden layer define the position and width of a radial basis function, and each linear output neuron forms a weighted sum of these radial basis functions. Figure 1 shows a schematic of the RBFN (Hagan et al., 1996).

The purpose of RBFN is to produce a generalized function by mapping the input to a set of high dimensional hidden functions, such that for a set of $X$ data points in a $D$-dimensional input, $X^P = \{X_i^P; i = 1,2,....,D\}$ to be mapped into the corresponding target, $t^p$ the RBFN is needed to obtain a generalized function $f(X^P)$ such that:

$$f(X^P) = t^p \qquad \forall p = 1,......,J \tag{5}$$

The RBFN method introduces a set of $J$ non-linear Gaussian functions for each neuron in the hidden layer, which take the form:

$$\varphi_l(x) = \exp(-\frac{\|x - \mu_l\|^2}{2\sigma_l^2}) \tag{6}$$

Where $\mu_l$ the basis is centre for the neurons, and $\sigma_l$ is the width of the basis function, determined by the training algorithm. Each of the approximated outputs has a linear activation function expressed as:

$$y_z(x) = \sum_{l=0}^{m} w_{zl}\varphi_l(x) \tag{7}$$

Where $w_{zl}$ is the synaptic weight connecting neuron $l$ in the hidden layer to neuron $z$ in the output layer, and whose value is determined by the network.

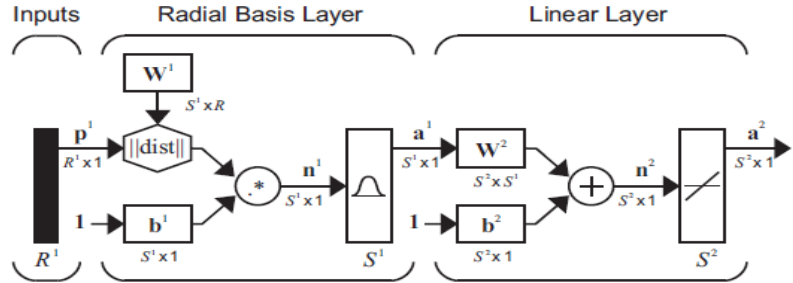

Figure 1: RBFN Design (Hagan et al., 1996).

RBFN is relatively faster than other multilayer perceptron and the computational requirements are relatively low. In this work, a Radial basis network is used for function approximation/regression tasks. Fig. 2 below shows the structure of fault identification and condition monitoring system.

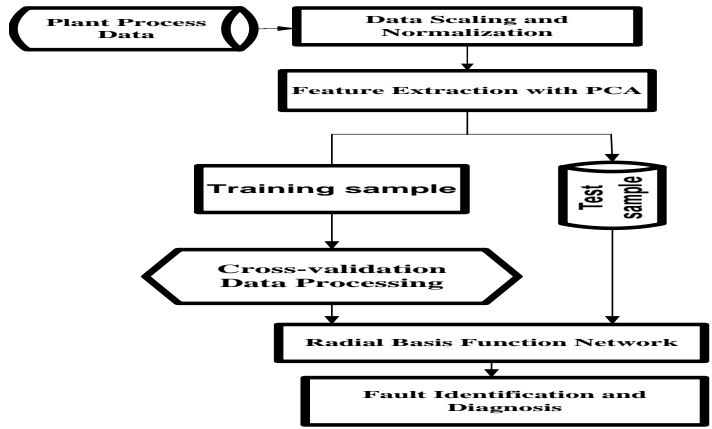

Fig. 2: A structural representation of the fault identification system

## 3. The Diagnosis process

### 3.1 The description of the simulated plant and the desktop simulator

Qinshan stage one phase II NPP is a Chinese version of the pressurized water reactor with a total (gross) capacity of 650MW. The Qinshan II NPP is a two-loop PWR that is based on the improved design of Qinshan I NPP. To simulate the plant and to obtain its representative data, the MST's Advanced Personal

Computer Transient Analyzer (PCTRAN) for PWR was utilized. The PCTRAN is a mimic of a two-loop PWR with the net thermal power of 1800 MW (Po, 2009). The purpose of this desktop simulator is to test the methodology on the simulation data from Qinshan II NPP that serves as a close representative of the dynamics common to the nuclear plant. Table 1 show a few selected initial conditions used as the normal operating parameters for the Qinshan II NPP simulation.

**Table 1** Plant parameters used for the Qinshan II NPP simulation.

| SYSTEM PARAMETERS | VALUE |
| --- | --- |
| Core thermal power | 1930MWt |
| Primary loop pressure | 158kg/cm$^3$ (155bar) |
| Average temperature of the coolant | 310$^o$C |
| Coolant loop flow | 46640m$^3$/h |
| Total core flow rate | 30530t/hr |
| Initial Pressurizer water level (FoF) | 0.565bar |
| HPI auto start set point(bar) | 129.69bar |
| Charging flow | 30t/hr |

### 3.2 Description of the malfunctions investigated

Different incipient faults were modeled on the simulator. An appropriate output pattern has been used to properly represent the faults. The sample output patterns for the six simulated fault cases are shown in Table 2. In the output pattern, the first digit represents the break size and the second digit represents the location of the fault. With the exception of 60% tube rupture fault in the Steam Generator B (SG-B), all other faults modeled in this work are the ones that occur without the activation of the Emergency Core Cooling Systems (ECCS), in the first 1000s. The time scale for each transient period is selected to be 1000 s. Table 2 below shows the malfunctions simulated, and the output variable of interest.

**Table 2** Malfunctions investigated and the expected outputs.

| Data Set | Simulated faults | Fault Size(%) and Location |
| --- | --- | --- |
| 1 | Normal operation | 0.00, 0.00 |
| 2 | 15% tube rupture faults in the SG-A | 15.0, 1.00 |
| 3 | 30% tube rupture faults in SG B | 30.00, 2.00 |
| 4 | 45% tube rupture faults in SG A | 45.00, 1.00 |
| 5 | 60% tube rupture fault on SG B with ECCS | 60.00, 2.00 |
| 6 | Locked rotor on the #1 RCS pump | 100.00, 3.00 |

### 3.2.1 Locked Rotor fault
A locked rotor event is characterized by the instantaneous seizure of a reactor coolant pump rotor. Flow through the affected reactor coolant loop is rapidly reduced, leading to a reactor trip on a low flow signal. Before the trip, there is a rapid power and pressure increase. A zero moderator temperature coefficient was assumed in the analysis (Po, 2009). A 100% fully locked rotor fault is simulated in the pump #1 of the RCS.

### 3.1 Parameter selection
For the faults analyses, Table 3 shows the selected important process parameters that showed significant variations. These parameters are derived when all plant data are normalized and the constant rows are removed. Subsequently, the parameters are used as the inputs to the PCA. The table consists of 43 out of over 80 parameters with time-dependent transient data representing the dynamics of the plant. All the faults are modeled using the two recurrent networks and the performance of each of the networks is as discussed in the next section.

**Table 3**: List of process parameters for fault analysis

| SI. No. | Parameter Label | Description | SI. No. | Parameter Label | Description |
|---|---|---|---|---|---|
| 1 | P | Pressure of RCS | 16 | TAVG | RCS Tempt (°C) |
| 2 | TCA,TCB | Temperature of Cold leg A and B | 17 | PSGA,B | Pressure Steam generator A,B (kg/cm2) |
| 3 | QMWT | Total Thermal Power | 18 | WSPY | Flow Pressurizer Spray (t/hr) |
| 4 | QMGA,B | Power of SG-A,B heat removal | 19 | TFSB | Temp Submerged fuel average (°C) |
| 5 | NSGA, NSGB | Narrow range level of SG-A,B (t/hr) | 20 | HTR | Power Pressurizer Heater (MW) |
| 6 | WFWA,B | Flow SG-A,B feedwater (t/hr) | 21 | PWR | Power Core thermal (%) |
| 7 | VOL | Volume RCS liquid (M3) | 22 | RBLK | Mass total leakage out of the Reactor Building (kg) |
| 8 | WRCA,B | Flow Reactor coolant loop A,B (kt/hr) | 23 | DNBR | Ratio departure from nucleate boiling |
| 9 | WSTA,B | Steam flow of SG-A,B | 23 | RHFL | Reactivity fuel %dk/k |
| 10 | LSGA,B | Level SG-A,B wide range (M) | 24 | RHMT | Reactivity moderator temperature (%dk/k) |
| 11 | VOID | Void of RCS (%) | 25 | RHRD | Reactivity Rod (%dk/k) |
| 12 | WEC | Flow Total ECCS (t/hr) | 26 | HUP | Specific enthalpy pressurizer top discharge (KJ/Kg) |
| 13 | LVPZ | Level Pressurizer (%) | 27 | HLW | Specific Enthalpy RCS leak (kJ/kg) |
| 14 | WTRA,B | Flow SG-A,B tube leak (t/hr) | 28 | TF | Average fuel Temperature (oC) |
| 15 | TF | Temp Average fuel (°C) | 29 | THA,THB, TCA,TCB | Temperatures of Hot Leg A, Cold leg A, hot leg B, cold leg B. respectively (oC) |

**4. Event description and simulation analysis.**

The first fault considered is 15% and 45% of the whole tube rupture in SG-A, and 30% and 60% rupture faults in SG-B. When an event occurs starting from the steady-state operation, instruments' readings develop a time- dependent pattern and these patterns are unique with respect to the type of an event (Santosh et al., 2009). That is, the variations in the values of system parameters are different as the magnitude of faults change. Hence, an estimation of the magnitude of a fault can be obtained by tracking the variation in operational parameters, as shown by the result of analyzing the simulation data. Some parameters show a step change after a certain fault occur, and show different variation as the degree of the fault changes. The following patterns are observed when rupture faults of different sizes were injected in the simulations: First, steady-state, no fault condition of the plant was implemented. After operating the plant under steady-state, normal condition for 50s, each of the faults was injected, and their corresponding parameters were observed for 1000 time steps. It was observed that there is a correlation between the deviation in parameters and the severity of the fault. That is the plant parameter deviation from normal increases as the severity of fault increases. Fig. 3 shows how the changes in the steam generator A heat removal capacity as the severity of the fault increases. Similarly, Fig. 4 shows the variation of a system parameter (feed water flow) as a function of the variation in fault severity in the system. Another example of similar analysis is as shown in Fig. 5.

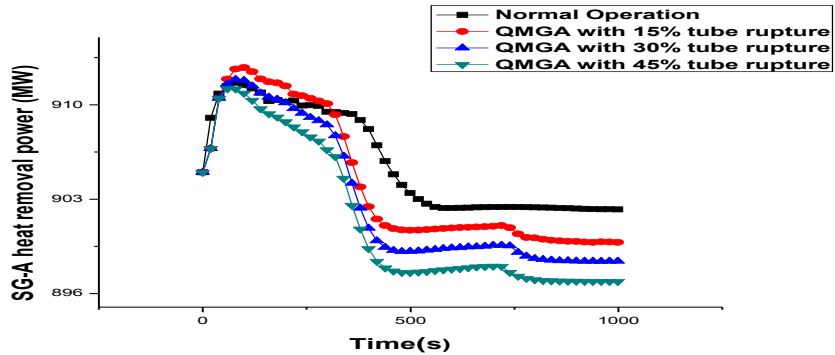

Fig. 3: Change in SG-A heat removal power with variation in tube rupture size

Similarly, fig 4 represents the deviation in SG-A feed water flow as the sizes of the tube rupture increases and figure 5 shows the change in the reactor coolant system inventory as the rupture fault increases.

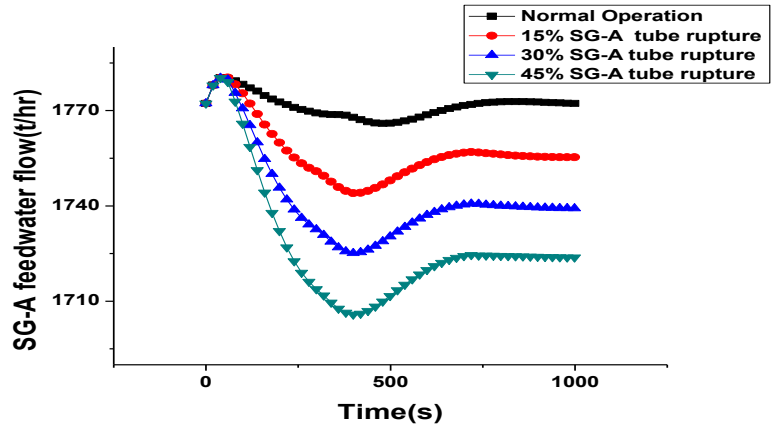

Fig 4: Deviation in SG-A feed water flow as the sizes of the tube rupture increases.

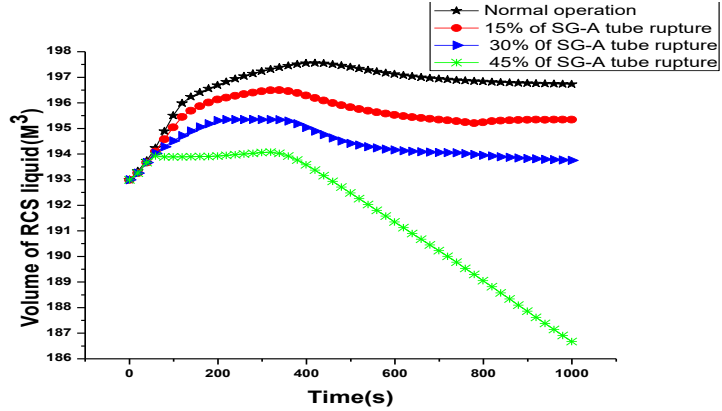

Fig. 5: Changes in the reactor coolant system inventory as the rupture fault increases.

Hence by properly reviewing the plant process parameters, faults could be detected and this pattern – based fault detection can serve as a form of pre-diagnosis steps.

### 4.1 RBFN training and results

In this section, we present the results obtained during experimental training of the network using different topologies and their generalization capability on different training sets. The training and testing of the neural networks are implemented using the MATLAB Machine learning toolbox. A total data set of 5446-time steps is partitioned into 3 different sets: Training set (70%), a Validation set (15%) and Test set (15%). The training set, containing 3814 vectors is used for training, and the validation and test sets containing 816 samples each is used to check the generalization ability of the networks. In the experimental design phase, the networks are trained adaptively. First, the input vectors were normalized to the zero mean and the unit standard deviation. In turn, the target output is also transformed taking into consideration the response range of the output neurons, and the maximum and minimum values of the training sequence were memorized and applied to the validation and test data. Moreover, to obtain the optimal training sets, we utilized PCA to filter the noise, select appropriate features and reduce the dimension of the plant parameters. That is, PCA is applied to all the plant parameters that showed significant deviations during the plant operations, as shown in Table 3, leaving only the components with above 2% change in value. After the PCA stage, the data is sequentially introduced into the networks.

### 5. Fault identification result

#### 5.1 Performance measure
Generally, the performance of a network is measured by the learning curve, which shows the prediction accuracy of the network on the test set as a function of the training set or the desired output as a function of the input vectors. In this work, the mean-square error (MSE), regression plots and the predicted values are the performance indicators to validate the performance of the evaluated models. To evaluate the generalization ability of the networks, the validation set and test sets are applied. In addition, an unlabelled fault of 40% tube rupture in the SG-B of the PWR is applied.

#### 5.3 training results for RBFN
To find the best performing model, we experimented with many network configurations. This is necessary to identify a model that could fit the simulated data set in the best way. A new RBFN was designed with the Mean Square Error (MSE) specified as 0.04, and the spread of the function is specified as 1. Since RBFN adds neurons to the hidden layer until the mean square error is met, the best network architecture is derived when the RBFN is trained in 1.27s after 300 epochs with 311 hidden neurons in the hidden layer, and the performance curve is as shown in fig. 10. The regression plots showing the performances on training, test and validation data for 311 hidden neurons structure of the RBFN are presented in fig. 11, 12, and 13 respectively.

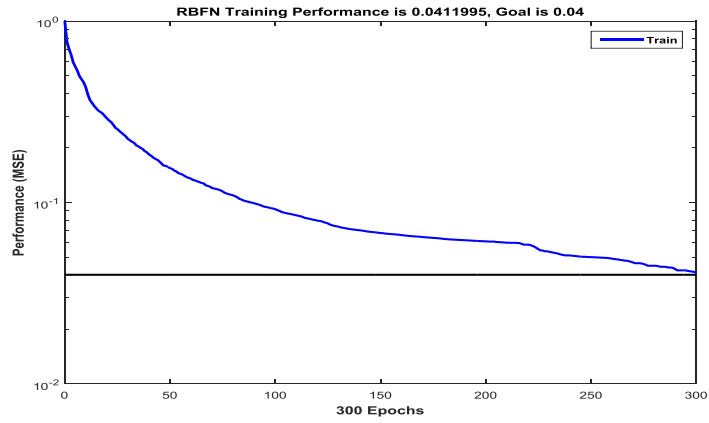

Fig.10: Training performance for RBFN

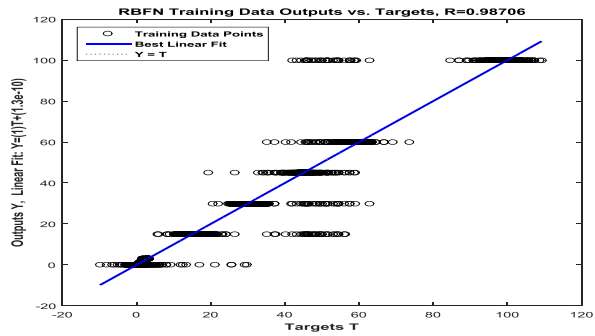

Fig 11: RBFN regression plot for the training data set.

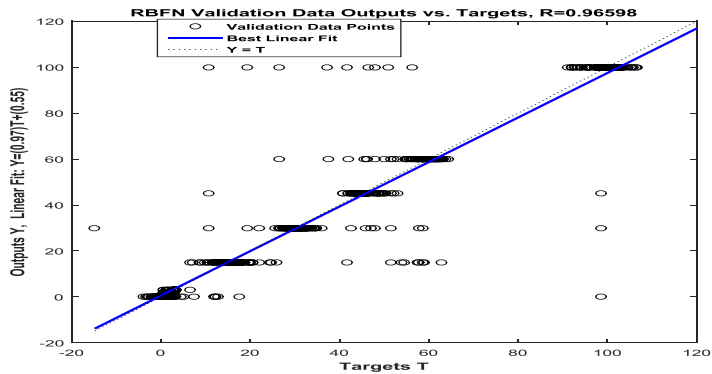

Fig 12: RBFN regression plot for the validation data set.

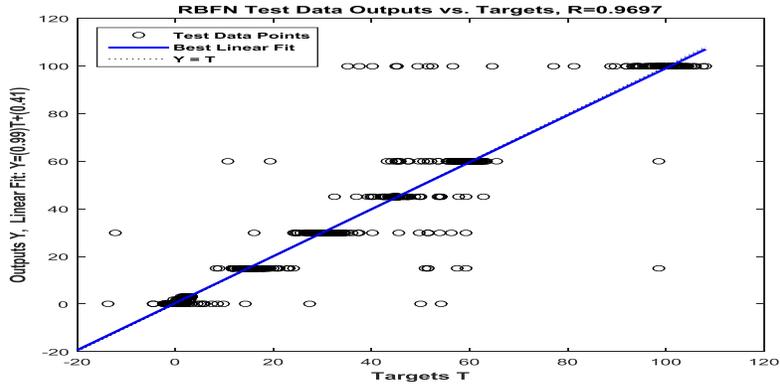

Fig 13: RBFN regression plot for the test data set.

### 5.1 RBFN evaluation

As observed on the RBFN regression plots, the average regression accuracy on the three data sets is 0.974, and the training time for RBFN is 1.27s over 300 epochs. Moreover, the best training accuracy obtained from the 311 neuron RBFN is 0.04. Although RBFN with 311 neurons in the hidden layer gives a more complex network, it compensated for the size of the network with the speed and accuracy of the results. To further check the quality of the results, the neural network model was tested using another data set of 1000 unlabelled samples. These samples are the simulation result of 40% and 50% tube rupture faults in the SG-A and SG-B of the PWR respectively. Table 5 and 6 show the values derived when the RBFN was tested on new unlabelled data and Fig 14 shows the regression plot of RBFN for the 40% tube rupture fault in SG A. As can be seen, the generalization of the RBFN dynamic network is quite good.

**Table 6 Result of RBFN on unlabelled data**

| Fault type | Target output | RBFN output (Avg) | RMS Error |
|---|---|---|---|
| 40% tube rupture fault in SG A | 40.0  1.00 | 40.0  0.99 | 0.007 |
| 40% tube rupture fault in SG B | 40.0  2.00 | 40.0  2.0 | 0.0 |
| 40% tube rupture fault in SG A | 50.0  1.00 | 49.9 1.0 | 0.071 |
| 40% tube rupture fault in SG B | 50.0  2.00 | 50.10  1.98 | 0.072 |

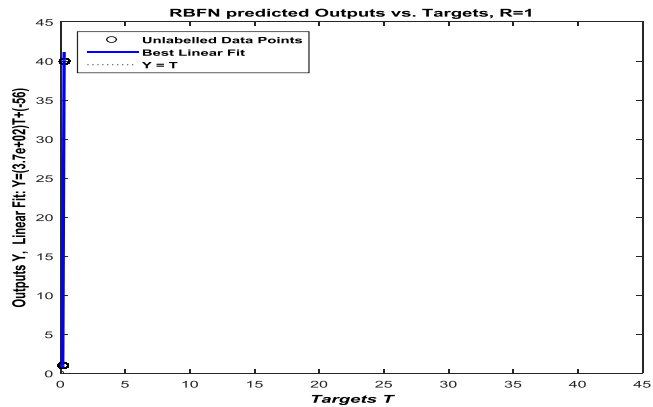

Fig 14: RBFN regression output for 40% rupture fault on SG-A

## 6. Discussion and Conclusion

A high-tech, high-performance system such as the nuclear power plant needs a wide range of support for operators to efficiently operate the plant, interpret and manage the volume of information available, and detect and diagnose a fault in a timely manner. In an effort to conduct a pilot study of different dynamic neural network architectures suitable for predicting and diagnosing faults specifically in components of complex systems like NPP, a few scenarios of PWR transients have been carried out, to generate simulated data corresponding to various occurrences of anomalies such as rupture faults in the steam generator and pump failure in the primary loop of Qinshan II NPP. Different rupture sizes ranging from 15%, 30% and 45% without ECCS activation, 60% with ECCS activation and a locked rotor fault without ECCS activation have been analyzed using ANNs. For the purpose of noise filtering, PCA was applied as a pre-diagnostic processing and data normalization phase. The ANN and PCA method was implemented with the MATLAB neural network toolbox. In order to have an idea about the generalization capability and suitability of a specific neural network for this regression task, and to compare the results from different network models, a number of recurrent RBF neural network architectures were tested on the representative data derived from a desktop simulator. The mean-square error (MSE), regression plots and predicted values are the performance indicators to validate the performance of the evaluated models. It was observed from this study that out of the network architectures tested, RBFN with 311 hidden neurons in the hidden layer shows the best performance, and it was adopted for further development of a knowledge base for the operator support system of Qinshan II NPP.

## 7. Acknowledgment

This research work was funded by the Natural Science Foundation of Heilongjiang Province, China (Grant NO.A2016002 and NO.E2015053), and the National Natural Science Foundation of China (NO.51379046).